\documentclass[conference]{IEEEtran}
\usepackage{cite}
\usepackage{amsmath,amssymb,amsfonts}
\usepackage{algorithmic}
\usepackage{graphicx}
\usepackage{textcomp}
\usepackage{xcolor}

\usepackage{listings}
\definecolor{myblue}{HTML}{336699}
\usepackage{marvosym}
\usepackage{tikz}
\usetikzlibrary{arrows, arrows.meta, calc, positioning, shapes}
\usepackage{multirow}
\usepackage{array}
\usepackage{arydshln}
\usepackage{placeins}
\usepackage{lipsum}
\usepackage[group-separator={,}]{siunitx}

\usepackage{url}

\def\BibTeX{{\rm B\kern-.05em{\sc i\kern-.025em b}\kern-.08em
    T\kern-.1667em\lower.7ex\hbox{E}\kern-.125emX}}
\begin{document}

\title{Faster and Simpler SNN Simulation\\with Work Queues}

\author{\IEEEauthorblockN{Dennis Bautembach}
\IEEEauthorblockA{\textit{FORTH\,-\,ICS} and \textit{CSD\,-\,UOC} \\
denniskb@ics.forth.gr
}
\and
\IEEEauthorblockN{Iason Oikonomidis}
\IEEEauthorblockA{\textit{FORTH\,-\,ICS} \\
oikonom@ics.forth.gr
}
\and
\IEEEauthorblockN{Nikolaos Kyriazis}
\IEEEauthorblockA{\textit{FORTH\,-\,ICS} \\
kyriazis@ics.forth.gr
}
\and
\IEEEauthorblockN{Antonis Argyros}
\IEEEauthorblockA{\textit{FORTH\,-\,ICS} and \textit{CSD\,-\,UOC} \\
argyros@ics.forth.gr
}

}

\maketitle

\begin{abstract}
We present a clock-driven Spiking Neural Network simulator which is up to 3x faster than the state of the art while, at the same time, being more general \textit{and} requiring less programming effort on both the user's and maintainer's side. This is made possible by designing our pipeline around ``work queues'' which act as interfaces between stages and greatly reduce implementation complexity. 
We evaluate our work using three well-established SNN models on a series of benchmarks.
\end{abstract}


\newcommand{\Neur}{N} 
\newcommand{\Syn}{Syn} 
\newcommand{\Sp}{S} 

\section{Introduction}
Spiking neural networks (SNNs) are an important class of artificial neural networks (ANNs) ~\cite{Maass1997b}. Compared to other types of ANNs, SNNs more closely mimic the function of biological neural networks, including human brains. One key difference between SNNs and other types of ANNs is the network topology. ANNs are typically modeled as multipartite graphs whereas SNNs are modeled as directed graphs. Another key difference is in the encoding of network activity. ANNs produce continuous outputs whereas SNNs emit so called ``spikes'' at distinct points in time.
The output of a SNN is its firing pattern over time. 

SNNs are fundamentally more powerful than first (perceptron, unidirectional, hand-crafted features, discrete output) and second generation ANNs (deep, bidirectional, continuous, trained automatically)~\cite{Maass1996}. They (re-)gained a lot of traction and popularity in recent years, promising to aid in understanding the function of human brains, to be better suited at processing spatio-temporal data (real-world sensory data), and to potentially outperform Deep Learning and ``classical'' (feed-forward) neural networks, which may reach a saturation point.

Nevertheless, their training and simulation remain unsolved problems. Training is made difficult because of the discontinuous nature of spikes, which prevents us from applying Backpropagation natively (adaptions of Backpropagation exist which approximate spikes with (sharp) continuous functions~\cite{tavanaei2018deep}). On the simulation side, the communication between neurons is the main bottleneck due to neurons being connected randomly (even when those connections are localized). This makes SNN simulation equivalent to the signal propagation problem in directed graphs. In contrast, the regular nature of conventional ANNs allows us to reduce neuron communication to highly efficient matrix multiplications.

We are reminded that GPU acceleration was a major contributor to the Deep Learning revolution, as it enabled training of large-scale ANNs ``at home''. A similar revolution has yet to happen in the SNNs domain. Towards this end, we present a clock-driven simulator for spiking neural networks, which improves upon the state of the art in terms of performance, generality, and ease of use. It scales to larger networks than our competition and its unique API allows the creation of custom models with minimal programming effort.
The simulator is publicly available at \url{https://github.com/denniskb/ijcnn2020}.
\begin{figure*}
    \newcommand{\CR}{2mm} 
    
    \tikzstyle{stage} = [minimum height = 3\baselineskip, rounded corners = 1.5mm, align = center, draw = black]
    \tikzstyle{arr} = [-{Classical TikZ Rightarrow}, rounded corners = \CR, thick, black!50]
    \tikzstyle{delay} = [cyan!50]
    \tikzstyle{plast} = [magenta!45]
    \tikzstyle{label} = [yshift = .5\baselineskip, font = \small, text = black!80]
    \tikzstyle{input} = [label, pos = 1, anchor = east]
    \tikzstyle{output} = [label, pos = 0, anchor = west]
    
    \centering
    \begin{tikzpicture}[x = 25mm, y = \baselineskip, node distance = 0]
        \matrix (pipeline) [column sep = 25mm, row sep = 0] {
            \node (nupdate) [stage] {Update\\Neurons}; & & &
            \node (receive) [stage] {Receive\\Spikes}; \\
            &
            \node (fifo) [align = center, draw] {FIFO\\$|delay{\color{magenta}+1}|$}; \\
            & &
            \node (supdate) [stage] {Update\\Synapses}; \\
        };
        \node [below = of pipeline, font = \footnotesize] {
            \begin{tabular}{llllll}
                {\color{black!50} \rule[2pt]{1em}{1.5pt}} & Core Pipeline \hspace{1em} & $N$ & Neurons \hspace{1em} & $t$ & Timestep \\
                {\color{cyan!50} \rule[2pt]{1em}{1.5pt}} & w/ Delays & $S$ & Spikes & $d$ & Delay \\
                {\color{magenta!50} \rule[2pt]{1em}{1.5pt}} & w/ Plasticity & $Syn$ & Synapses \\
            \end{tabular}
        };
 
        \foreach \n in {nupdate, fifo, supdate, receive} {
            \foreach \i in {1, ..., 3} {
                \coordinate (\n i\i) at ($(\n.west) + (0, 2\baselineskip -\i\baselineskip)$);
                \coordinate (\n o\i) at ($(\n.east) + (0, 2\baselineskip -\i\baselineskip)$);
            }
        }
        
        \coordinate (NIN) at ($(nupdate.west) + (-.33, 0)$);
        \draw [arr, -{Triangle}] (NIN) -- ++(1pt, 0);
        \draw [arr, -] (NIN) -- node[output]{$\Neur_t$} (nupdate);
        \draw [arr] (nupdateo1) -- node[output]{$\Neur_{t+1}'$} node[input]{$\Neur_{t+1}'$} (receivei1);
        \draw [arr, plast] (nupdateo2) ++(.35, 0) -- +(\CR, 0) |- node[input]{$\Sp_t$} (fifo);
        \draw [arr, plast] (nupdateo2) ++(.35, 0) -- +(\CR, 0) |- node[input]{$\Sp_t$} (supdatei2);
        \draw [arr, delay, loosely dashed, very thick, -] (nupdateo2) ++(.35, 0) -- +(\CR, 0) |- (fifo);
        
        \coordinate (NOUT) at ($(receiveo2) + (.4, 0)$);
        \draw [arr, -{Triangle}] (receiveo2) -- node[input]{$\Neur_{t+1}$} (NOUT);
        
        \coordinate (x) at ($(fifo.east |- receive) + (.45, 0) + (2*\CR, 0)$);
        \draw [arr, delay, -] (fifoo2) ++(.45, 0) -- ++(2mm, 0) |- (x);
        \draw [arr, plast] (fifoo2) -- node[output, yshift = -\baselineskip, text = magenta]{$\Sp_{t-d}$} ++($(.45, 0) + (\CR, 0)$)  |- node[input]{$\Sp_{t-d}$} (supdatei1);
        \draw [arr, delay, loosely dashed, very thick, -] (fifoo2) -- node[output, text = cyan]{$\Sp_{t-d+1}$} ++(.45, 0);
        \draw [arr] (nupdate) -- node[output]{$\Sp_t$} node[input]{$\Sp_{t-d+1}$} (receive);
        
        \draw [arr, plast] (supdateo3) -- node[output]{$\Syn_{t+1}$} ++(.5, 0) |- node[input]{$\Syn_{t+1}$} (receivei3);
        \draw [arr, plast, -{Triangle}] (supdateo3) -- node[input]{$\Syn_{t+1}$} (NOUT |- supdateo3);
        \draw [arr, plast, -{Triangle}] (NIN |- supdatei3) -- ++(1pt, 0);
        \draw [arr, plast] (NIN |- supdatei3) -- node[output]{$\Syn_t$} node[input]{$\Syn_t$} (supdatei3);
    \end{tikzpicture}
    \caption{A single loop iteration of our simulator's pipeline, transitioning the SNN from timestep $t$ to $t+1$. Optional, feature-dependent paths are color-coded. (not depicted) The "Init" stage initializes neurons and synapses to $\Neur_0$ and $\Syn_0$.}
    \label{figure:pipeline}
\end{figure*}
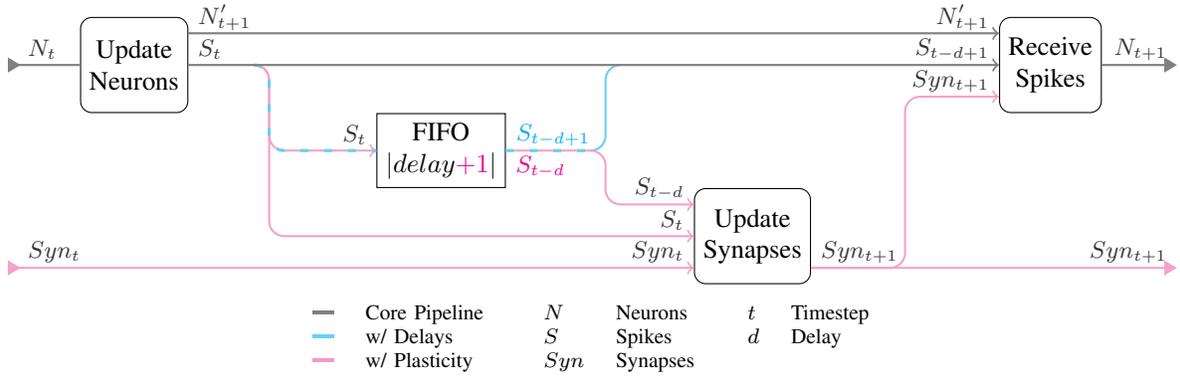

\section{Related Work}
Research related to the design and simulation of SNNs has a long history~\cite{McCulloch1943}, using both hardware~\cite{Tomlinson1990,Pelayo1997}, and software~\cite{Mattia2000,Reutimann2002}. Each approach has its own advantages and disadvantages, and distinct approach subclasses can be identified.

Older approaches have capitalized on advances in processing power~\cite{Mattia2000} and hardware design~\cite{Tomlinson1990}. A more recent trend follows the ubiquitous success of Deep Learning, which largely capitalizes on the advances in commodity GPU processing power~\cite{Yavuz2016,Ahmad2018,Thibeault2011,Fidjeland2010,Fernandez2008}.
In all software solutions to the problem, an important distinction can be made~\cite{Garrido2014}: event-driven vs clock-driven.
Event-driven methods adopt a mostly asynchronous approach, recording events and processing their implications in an on-demand manner. On the other hand, time- or clock-driven methods advance the whole network state in lockstep. Furber et al.~\cite{Furber2013} pursue a hybrid approach: their fundamental design revolves around an event queue but they also organize spikes into 100$\mu$\textit{s} long bins. All spikes from a bin are regarded as taking place simultaneously and thus processed in parallel. 

\subsection{Hardware}

Designing hardware specifically for the task of SNN simulation has a long history~\cite{Tomlinson1990,Pelayo1997,Schemmel2010} and is still an actively researched area~\cite{Furber2013,Sripad2018,Lee2018}. 
Custom-designed hardware in earlier attempts~\cite{Tomlinson1990,Pelayo1997,Schemmel2010} has been recently replaced by designing FPGA solutions~\cite{Sripad2018}, and partial components, which can be integrated into larger chips\cite{Lee2018}.
Finally, combinations of software and hardware components ~\cite{Ros2006a}, as well as very large scale, neuromorphic systems~\cite{Furber2013} have been presented.

\subsection{Simulators}

In parallel to dedicated hardware, a significant amount of effort has been devoted to developing general-purpose software simulators of SNNs.
Such general-purpose simulators include the NEST simulator~\cite{Eppler2009} and the Brian simulator~\cite{Goodman2010} among others~\cite{Pecevski2014}. Research on SNNs has recently focused on using software simulation~\cite{Rudolph2006,Ros2006,Morrison2007,Brette2007}.

\subsection{GPU}

Several recent software simulation approaches resort to GPU acceleration. Namely, the projects GeNN~\cite{Yavuz2016}, Brian2GeNN~\cite{Stimberg2018}, CARLsim~\cite{Chou2018}, SPIKE~\cite{Ahmad2018}, Fujita et al.~\cite{Fujita2018}, Kasap et al.~\cite{Kasap2018} are all recent (i.e. within the last four years) approaches that employ GPU acceleration.
Out of these, GeNN~\cite{Yavuz2016} and SPIKE~\cite{Ahmad2018} are two actively developed projects and, to the best of our knowledge, are two of the fastest GPU-accelerated SNN simulators, with different trade-offs. SPIKE is heavily optimized for simulation speed, and in doing so, sacrifices some generality and memory efficiency. On the other hand GeNN exposes a versatile programming interface to the user allowing them to implement custom models, at the cost of some speed. We therefore compare the performance of the proposed simulator against these two, as representatives of the state of the art that cover a wide spectrum of trade-offs.

A very recent trend in simulating SNNs is based on the use of the widely popular Deep Learning frameworks. More specifically, BindsNET~\cite{Hazan2018} and SpykeTorch~\cite{Mozafari2019} are both based on the PyTorch~\cite{paszke2017automatic} framework to simulate SNNs.
In practice, such approaches do not support general SNN simluation, limiting their applicability.

Overall, the presented SNN simulator leverages the power of consumer GPUs. We adopt a clock-driven approach, and use work-queues for a computationally efficient implementation.
\section{Methodology}
\subsection{Architecture}
\subsubsection{Pipeline}

We draw inspiration from~\cite{HLBVH2011} and organize our simulator into stages which communicate through ``work queues'': $Stage_i$ produces a queue and $Stage_{i+1}$ consumes it. The queues act as interfaces, decoupling stages, and allow to implement $k$ features with $O(kn)$ effort, $n$ being the number of stages, as opposed to $O(k^n)$ if the stages were tightly coupled. Our core simulation logic is only 70 lines of code long.

On a high level, our simulator consists of three stages: \lstinline{Init} initializes neurons $\Neur$ and synapses $\Syn$ to $\Neur_0$ and $\Syn_0$.
Then the simulation enters a cycle of \lstinline{Update Neurons}$\rightarrow$\lstinline{Receive Spikes} steps, advancing the SNN from timestep $t$ to $t+1$.
\lstinline{Update Neurons} advances neuron dynamics from $\Neur_t$ to $\Neur'_{t+1}$ and, in doing so, potentially generates a queue of spikes $\Sp_t$. \lstinline{Receive Spikes} consumes $\Neur'_{t+1}$ and $\Sp_t$, delivers the spikes to their recipients, and produces $\Neur_{t+1}$ (Fig.~ \ref{figure:pipeline}, black paths only).

This effectively implements non-instantaneous connections, which are necessary to simulate SNNs that contain cycles.
Non-instantaneous connections \textit{imply a synaptic delay of at least 1} timestep. An important feature are arbitrary delays, which allow the simulation of spikes with travel times. Delays are implemented by routing the spikes generated by \lstinline{Update Neurons} through a first in -- first out (FIFO) queue with delay $d$ many entries, \textit{delaying} their arrival at \lstinline{Receive Spikes} and thus their delivery to their recipients. \lstinline{Update Neurons} and \lstinline{Receive Spikes} themselves need not change, they simply write to/read from different queues (Fig.~ \ref{figure:pipeline}, black \& blue paths).

Another important feature is spike-timing dependent plasticity (STDP) \cite{taylor1973problem}, which can be used to implement Hebbian learning~\cite{kempter1999hebbian,kozdon2018evolution}--it is the ability of a synapse to modify its state (typically its weight) depending on local network activity (pre- and postsynaptic spikes). Plasticity is implemented by adding an \lstinline{Update Synapses} step to the pipeline. \lstinline{Update Synapses} advances synapse dynamics from $\Syn_t$ to $\Syn_{t+1}$, querying $\Sp_{t-d}$ and $\Sp_t$ to determine pre- and postsynaptic spikes respectively (in practice we store bitmasks in addition to queues to speed up those queries). Since we need access to both $\Sp_{t-d}$ and $\Sp_t$ simultaneously, the size of the FIFO queue is increased to $d+1$. \lstinline{Receive Spikes} now takes into account $\Syn_{t+1}$ when delivering spikes to their recipients (Fig.~ \ref{figure:pipeline}, black \& pink paths).

In our case we actually implement a lazy variation of plasticity. Rather than eagerly advancing synapse dynamics at every simulation step, we intentionally keep synapses in a stale state and only update them on a need-to-basis, namely if either of two conditions occur:
\begin{itemize}
    \item A synapse is about to transmit a spike.
    \item A synapse is about to ``expire'', i.e. its age is about to exceed the size of our FIFO queue, after which we would be unable to update the synapse because we would loose access to pre/post-synaptic spike information.
\end{itemize}
In either case, the synapse is repeatedly updated in a loop until its state is current again. While the amortized number of updates remains identical to the eager version, we still observe a 4x performance gain in practice because the updates are now performed inside registers, avoiding global memory traffic. By abstaining from requiring a closed-form solution for synapse dynamics, which would allow us to update them in a single step (as done, for example,  by SPIKE~\cite{Ahmad2018}), we stay general and continue to support models that do not have closed-form solutions for their synapse dynamics. Since all out-going synapses of a neuron transmit together, their ages always stay in sync, meaning it is sufficient to store a single age per neuron and synapses can be updated in batches. Lazy plasticity is implemented by letting \lstinline{Update Neurons} produce one additional queue of ``expiring neurons'' and \lstinline{Update Synapses} updating only those synapses belonging to currently spiking or expiring neurons. We also increase the size of our FIFO queue from $d + 1$ to 50 entries (determined empirically) to reduce the frequency of these updates.

\subsubsection{Data Structures}
Subsequently we describe the various data structures employed by our simulator. For brevity, statements regarding neurons also hold for synapses.

We use simple arrays to store most of the SNN state. Users communicate their neurons' fields to us via variadic templates (see section~\ref{chapter:api}), which we, using template meta programming, convert into a structure of arrays (SoA)--one array for each field. Except during SNN instantiation and adjacency list construction we have no notion of neuron populations. Instead, users have to create fat neuron layouts and fat callbacks\footnote{A structure whose fields are the union of fields of many structures is called fat. A function that incorporates many different code paths, often via a series of if else-switches, is called fat.}.
The advantage is that we can store all neurons in a single SoA, simplifying their traversal, cutting down on kernel invocations, and simplifying indexing into said arrays from the adjacency list.

Queues are simply arrays bundled with an atomic index residing in global memory (for insertions), which is plenty fast for low contention scenarios. They are sized conservatively to avoid re-allocations. Table~\ref{table:memory} shows a detailed breakdown of our simulator's memory consumption.

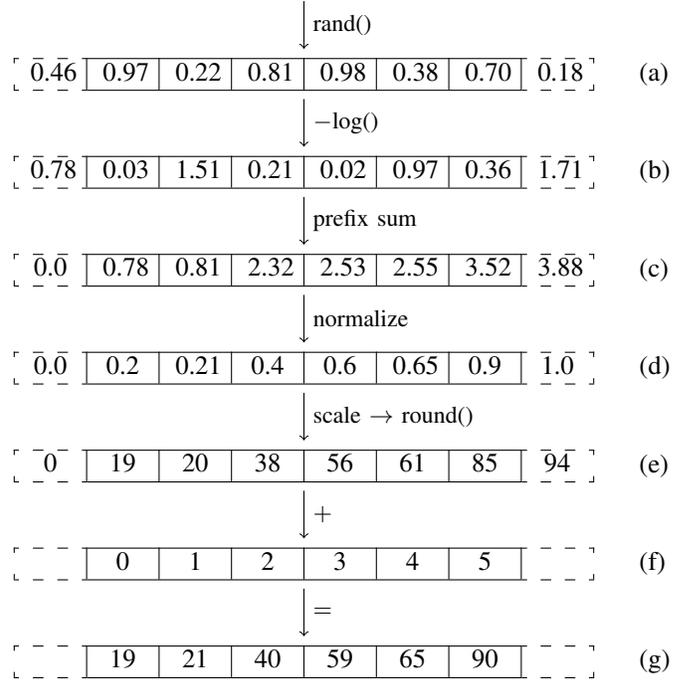
\begin{figure}
    \newcolumntype{P}[1]{>{\centering\arraybackslash}p{#1}}
    \newcommand{\step}[1]{
        \begin{tabular}{:P{1.5em}|P{1.5em}|P{1.5em}|P{1.5em}|P{1.5em}|P{1.5em}|P{1.5em}|P{1.5em}:}
            \cdashline{1-1} \cline{2-7} \cdashline{8-8}
            #1 \\
            \cdashline{1-1} \cline{2-7} \cdashline{8-8}
        \end{tabular}}
        
    \tikzstyle{arr} = [-{Classical TikZ Rightarrow}, thin]
    \tikzstyle{label} = [right, font = \small]
    
    \begin{tikzpicture}[node distance = 1.5\baselineskip and 1em]
        \node (origin) {};
        
        \node (a) [below = of origin] {\step{0.46 & 0.97 & 0.22 & 0.81 & 0.98 & 0.38 & 0.70 & 0.18}};
        \node [right = of a] {(a)};
        
        \node (b) [below = of a] {\step{0.78 & 0.03 & 1.51 & 0.21 & 0.02 & 0.97 & 0.36 & 1.71}};
        \node [right = of b] {(b)};
        
        \node (c) [below = of b] {\step{0.0 & 0.78 & 0.81 & 2.32 & 2.53 & 2.55 & 3.52 & 3.88}};
        \node [right = of c] {(c)};
        
        \node (d) [below = of c] {\step{0.0 & 0.2 & 0.21 & 0.4 & 0.6 & 0.65 & 0.9 & 1.0}};
        \node [right = of d] {(d)};
        
        \node (e) [below = of d] {\step{0 & 19 & 20 & 38 & 56 & 61 & 85 & 94}};
        \node [right = of e] {(e)};
        
        \node (f) [below = of e] {\step{\ & 0 & 1 & 2 & 3 & 4 & 5 & \ }};
        \node [right = of f] {(f)};
        
        \node (g) [below = of f] {\step{\ & 19 & 21 & 40 & 59 & 65 & 90 & \ }};
        \node [right = of g] {(g)};
        
        \draw [arr] (origin) -- node[label]{rand()} (a);
        \draw [arr] (a) -- node[label]{$-$log()} (b);
        \draw [arr] (b) -- node[label]{prefix sum} (c);
        \draw [arr] (c) -- node[label]{normalize} (d);
        \draw [arr] (d) -- node[label]{scale $\rightarrow$ round()} (e);
        \draw [arr] (e) -- node[label]{$+$} (f);
        \draw [arr] (f) -- node[label]{$=$} (g);
    \end{tikzpicture}
    \caption{Efficient generation of sorted, uniformly distributed random integers on the GPU.}
    \label{figure:adj}
\end{figure}

One of the more interesting data structures is our adjacency list. We use a padded 2D array with $|\Neur|$ rows similar to~\cite{Yavuz2016}. Each row stores indices to all its neuron's neighbors. Rows shorter than the maximum degree $deg_{max}$ are padded with sentinels. This introduces a memory overhead of a few percent, but in exchange makes index calculations trivial, cuts down on global memory accesses (for the offset table), and allows us to tune row alignment. In practice, we measure a performance gain of a few percent with each row aligned to 128 bytes compared to a compact adjacency list, which requires an additional offset table. In the case of models with synapse state we allocate $|\Neur|*deg_{max}$ synapses with an implicit 1:1 mapping between the adjacency list and the synapse SoA.

As for traversal, we launch one CUDA block~\cite{cuda} per spike, which reads the corresponding row from the adjacency list and delivers the spike to its recipients. We also experimented with launching one warp per spike and with launching one thread per recipient neuron (, which~\cite{Yavuz2016} refer to as ``postsynaptic parallelism''), both of which performed worse in our benchmarks. Delivering a spike results in poor memory access patterns since all neurons' neighbors are scattered across the whole neuron SoA due to random connectivity.
We try to alleviate this somewhat by sorting each row, improving cache locality.

Adjacency list construction is a two step-process. On the CPU, we consume the user-provided SNN description (number and sizes of neuron populations and their connectivity, see listing~\ref{listing:tutorial}, lines 31 -- 32) and generate a queue of jobs $\{(n, a, b, o), ...\}$. Each job can be read as: ``Write $n$ sorted, uniformly distributed random integers from the interval $[a, b)$ into the adjacency list starting at offset $o$''. The jobs are uploaded to and processed by the GPU. In order to efficiently generate sequences of sorted random numbers, we take advantage of the fact that the sum of exponentially distributed random numbers is uniformly distributed. Fig.~\ref{figure:adj} depicts how a job is expanded for $n=6$ and $[a, b) = [0, 100)$: First, we generate six uniformly distributed random numbers from the interval $(0, 1]$ (\ref{figure:adj}a). We also pad our buffer with two sentinels whose purpose will become apparent shortly. Next, we obtain exponentially distributed numbers by computing the negative (natural) logarithm of these numbers (\ref{figure:adj}b). Afterwards we compute their running sum (\ref{figure:adj}c). At this point we already have obtained a list of sorted, uniformly distributed random numbers. The subsequent steps merely serve to transform this list into the desired range $[a, b)$. We normalize our list (\ref{figure:adj}d), scale it by $b-n = 94$ (\ref{figure:adj}e), and add consecutive integers $\{0, ..., n-1\}$ to it (\ref{figure:adj}f -- \ref{figure:adj}g) in order to ensure that each number is unique (i.e. that the SNN contains only single edges). By using sentinels, the first and last elements of our final list \textit{can} be equal to $a$ and $b-1$, but need not to. Without them, the final list would always be of the form $\{a, ..., b-1\}$, which would introduce a non-uniform bias.

\begin{lstlisting}[label = listing:tutorial, caption = {A simple SNN implemented using our framework.}, captionpos = b, float]
struct ping_pong : model
{
  struct neuron : neuron_desc<bool>
  {
    template <class Iter>
    void init(Iter it)
    {
      if (it.id() < 100)
        it.get<0>() = true;
      else
        it.get<0>() = false;
    }

    template <class Iter>
    bool update(Iter it, float dt)
    {
      bool spike = it.get<0>();
      it.get<0>() = false;
      return spike;
    }

    template <class Iter>
    void receive(Iter from, Iter to)
    {
      to.get<0>() = true;
    }
  };
};

snn<ping_pong> net(
  {100, 100},
  {{0, 1, 0.01}, {1, 0, 0.01}},
  1,
  1
);

while (true)
  net.step();
\end{lstlisting}

In practice, steps (a) -- (c) and (d) -- (g) can be executed in a single pass, respectively. The first pass is performed inside shared memory while the second pass only writes the final list to global memory. Random numbers are generated directly inside the kernel using the Xorshift RNG~\cite{xorshift2003}. Since the jobs have varying lengths we assign one warp per job. Warps can be scheduled independently from one another and thus do not hold each other hostage like a long-running thread would a block for example. Furthermore, threads of the same warp can communicate cheaply among themselves via warp-level primitives, making prefix sum calculations very fast. Finally, the CPU part's cost is completely hidden in practice because it is performed in the background
of GPU computations.
Our adjacency list construction algorithm achieves 90\% of the GPU's maximum memory bandwidth.

\subsection{API}
\label{chapter:programming_model}
Our superior performance (see section~\ref{chapter:results}) does not come at the cost of either generality or usability. On the contrary, we support a wide variety of SNNs by allowing users to implement their own, custom models with minimal programming effort. We present and discuss our API, followed by a comparison with SPIKE and GeNN.

\subsubsection{API}
\label{chapter:api}
We draw inspiration from modern graphics pipe\-lines such as DirectX and OpenGL: The GPU facilitates efficient rasterization, texel interpolation, texture filtering, etc, while allowing the user to customize the appearance of the final image through programmable shaders. Similarly, our simulator facilitates efficient spike propagation across the network, handling of delays, plasticity, etc, while allowing the user to customize the behavior of their model by invoking user-defined callbacks.

Let us walk through the implementation of a simple SNN with two randomly inter-connected neuron populations $A$ and $B$ of 100 neurons each that take turns exciting one another (listing~\ref{listing:tutorial}).
We begin by declaring a struct with our model name and inheriting from the ``model'' interface (line 1).
Next we add a child struct ``neuron'' to our model and communicate our neuron's layout by inheriting from ``neuron\_desc''. In our case neurons have a single field of type bool.
Next we have to implement a series of callbacks to define our model's behavior. We start by implementing ``init()'', which will be called once before the simulation and initialize the first neuron population to true, the second one to false. We get access to our neuron through an iterator (lines 5 -- 12).
Next we implement ``update()'', which will be called on every simulation step. If we previously received a spike, we emit one and reset ourselves so we do not spike again until we receive another one (lines 14 -- 20).
Lastly, we implement ``receive()'', which will be called whenever we receive a spike from another neuron. If that happens, we simply set our flag to true, which will cause us to spike during the next update step (lines 22 -- 26). 
Now that our model is defined, we can instantiate a SNN with it (line 30), passing neuron populations count and sizes (line 31), neuron populations connectivity (line 32), timestep (line 33), and delay (line 34) --- and run our simulation (lines 37 -- 38). 

The advantages of our approach are:
\begin{itemize}
    \item \textbf{Generality}: Any model that can be expressed using (a) the network state information provided by the framework\footnote{Currently the framework provides the user with information about neuron and synapse states and types, local connectivity, neuron/synapse populations count and sizes, etc., allowing the implementation of a variety of models. In theory, the network's total state could be exposed to the user, but this would defeat the purpose of a framework in the first place, whose role is to strike a balance between enabling users while relieving them at the same time.} and (b) the \lstinline{Update Neurons}$\rightarrow$\lstinline{Receive Spikes}-simulation loop, can be implemented. The implementation can use any CUDA C features
    or third party libraries. With this flexibility also comes responsibility:
    For example, the user has to (remember to) use atomic operations for updating neurons. Such intricacies can mostly be avoided by using existing building blocks, but become necessary when implementing esoteric models ex nihilo.

    \item \textbf{Composability and Reusability}: Class inheritance with method specialization leads itself to composability. Models need not be authored ex nihilo every time, but common components (such as leaky integrate and fire (LIF) neurons) can be extracted into their own classes and reused. Composability in turn increases reusability of such building blocks: If an existing building block does not meet the user's requirements it can be extended and \textit{some} of its functionality specialized, most of it reused. A great example of this approach can be seen in Fig.~~\ref{figure:loc_comparison}d: We implement the Brunel model (, which consists of Poisson and LIF neurons) by inheriting from the provided LIF neuron type and specializing its update method. Inside we spike randomly if we are a Poisson neuron, and simply delegate the call to the parent method otherwise.
    
    \item \textbf{Transparency}: The code written ends up being compiled by the native toolchain, making it easy to build a mental model.
    The compiler is able to provide detailed warnings and error messages. The code can be statically analyzed, debugged, and profiled (e.g. using NVIDIA Nsight).
\end{itemize}

\subsubsection{Comparison with SPIKE and GeNN}
SPIKE and GeNN employ different means to enable the composition and simulation of SNNs. SPIKE is a runtime library. It ships with a vast collection of popular, highly parameterizable neuron and synapse models that can be used as building blocks. Fixing the building blocks in place lends itself to relentless optimization. However, this comes at the expense of generality: Models that cannot be expressed as a combination of said building blocks, cannot be simulated using SPIKE, unless the authors release new building blocks on-demand. Implicit initialization is performed on the CPU, serially, and with quadratic complexity in the number of synapses. This made it impractical for our experimentation on large networks. To circumvent this and make experiments tractable, we employed user-level multi-threaded (CPU) explicit initialization, with linear complexity. Explicit initialization on the GPU would have been significantly faster, but SPIKE does not provide this option and its API cannot readily accommodate user-level provision. Simulation is performed entirely on the GPU, with the option to ``download'' timestamped neuron spikes.

\begin{figure}
    \begin{minipage}[t]{0.24\linewidth}
        \includegraphics[page = 1, width = \linewidth, trim = {34mm 30mm 60mm 32mm}, clip]{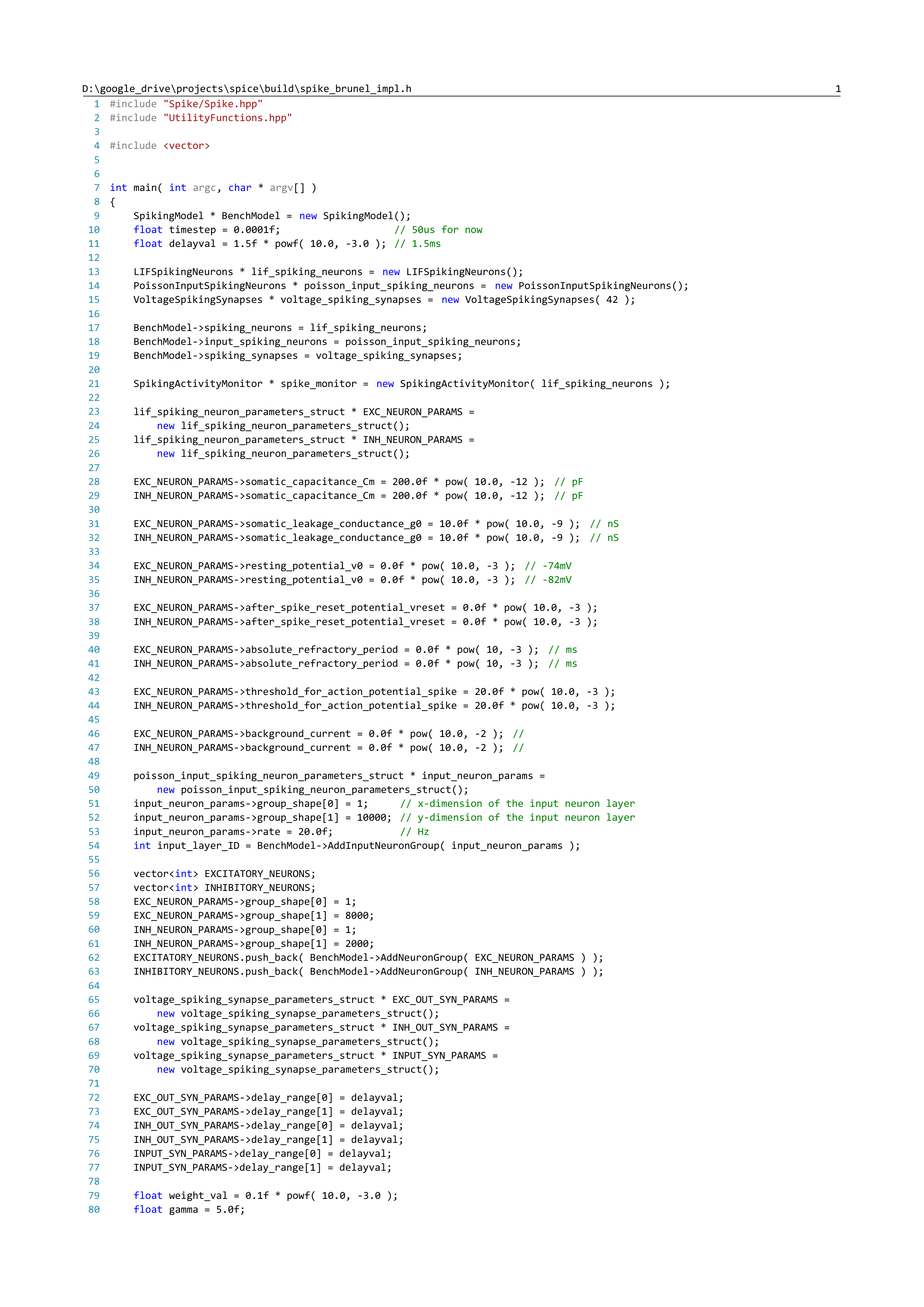}
        \includegraphics[page = 2, width = \linewidth, trim = {34mm 30mm 60mm 32mm}, clip]{figures/loc_comparison/spike.pdf}
    \end{minipage}
    \begin{minipage}[t]{0.24\linewidth}
        \includegraphics[page = 1, width = \linewidth, trim = {34mm 30mm 60mm 32mm}, clip]{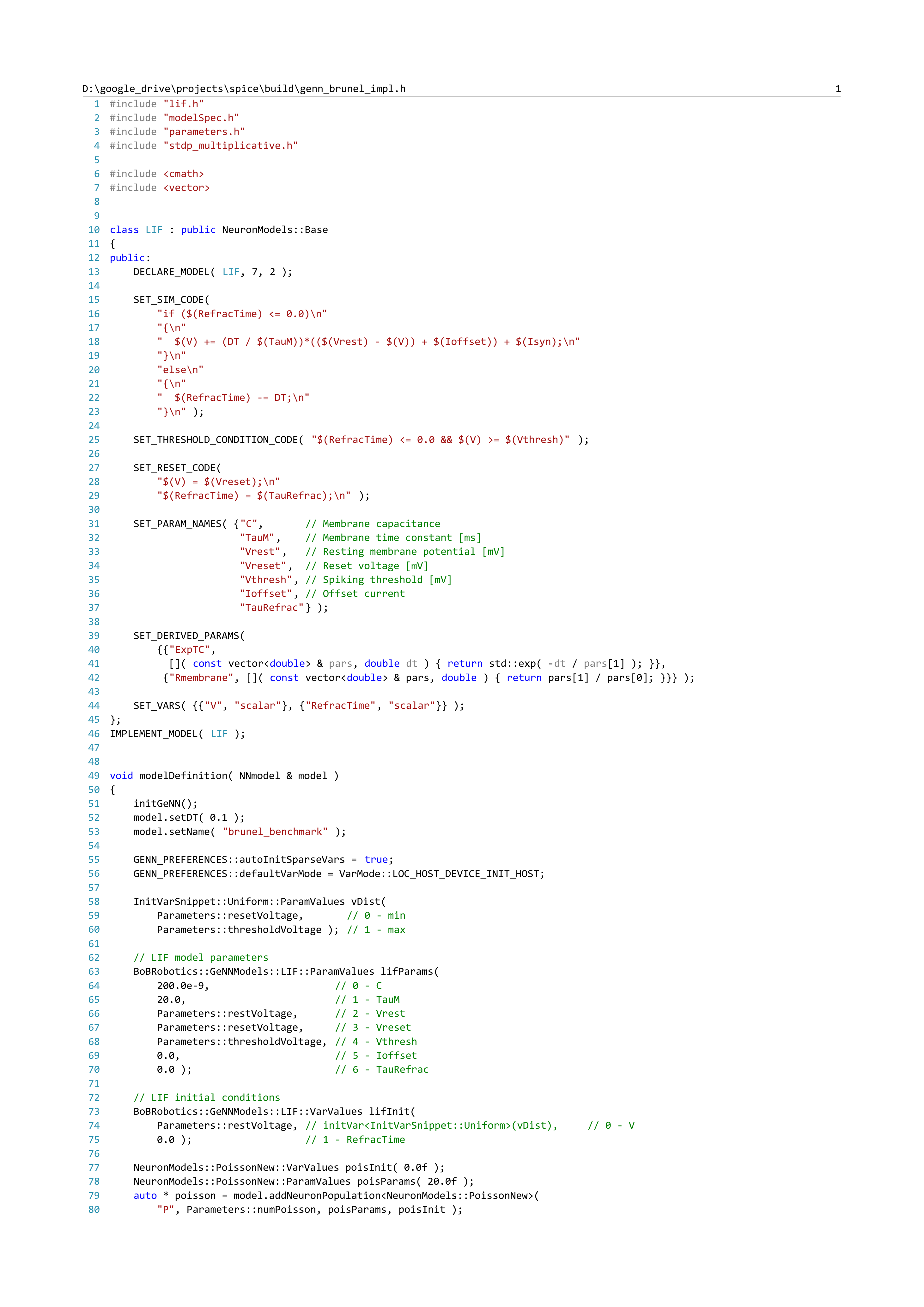}
        \includegraphics[page = 2, width = \linewidth, trim = {34mm 30mm 60mm 32mm}, clip]{figures/loc_comparison/genn.pdf}
        \includegraphics[page = 3, width = \linewidth, trim = {34mm 100mm 60mm 32mm}, clip]{figures/loc_comparison/genn.pdf}
    \end{minipage}
    \includegraphics[width = 0.24\linewidth, trim = {34mm 30mm 60mm 32mm}, clip]{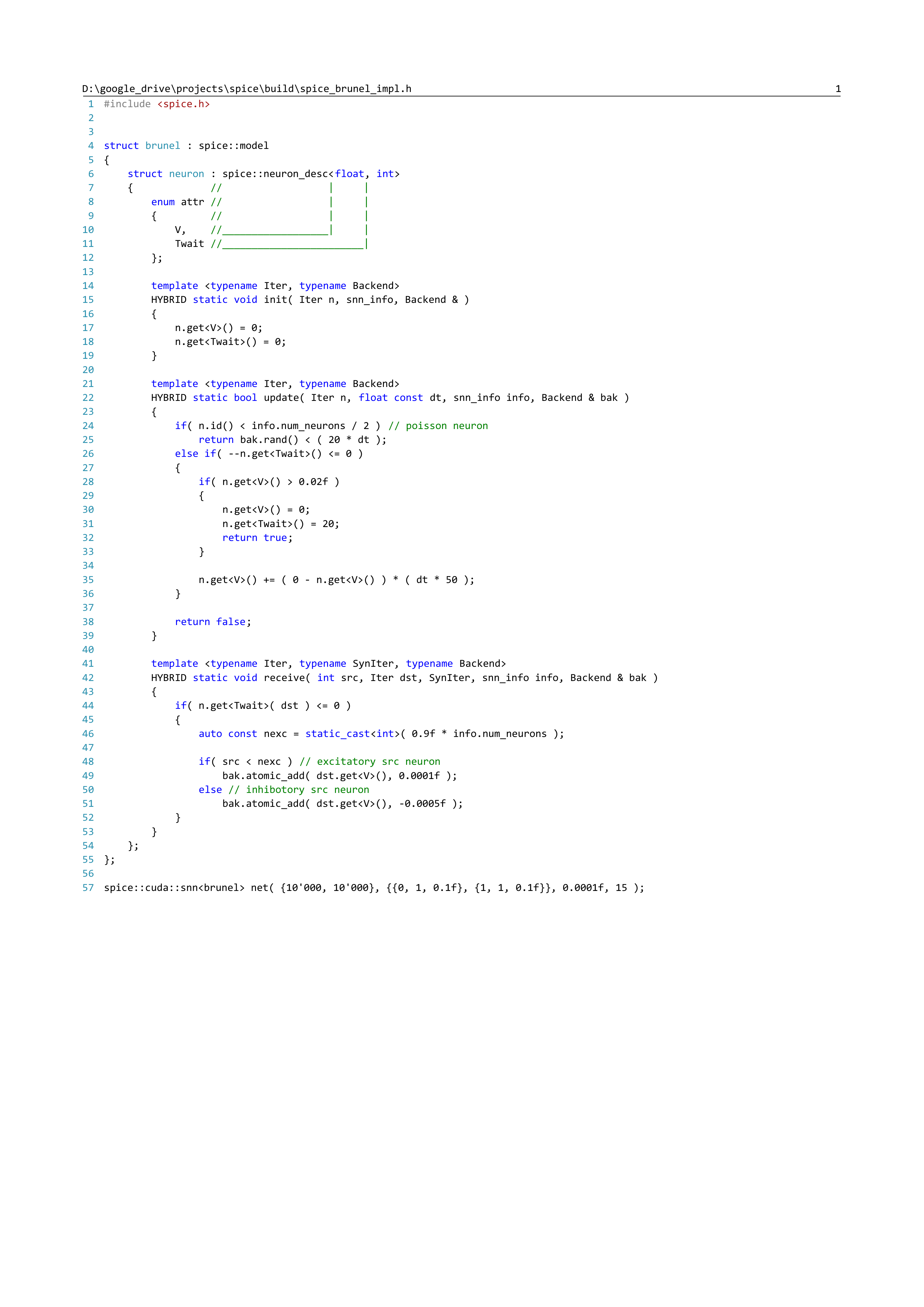}
    \includegraphics[width = 0.24\linewidth, trim = {34mm 30mm 60mm 32mm}, clip]{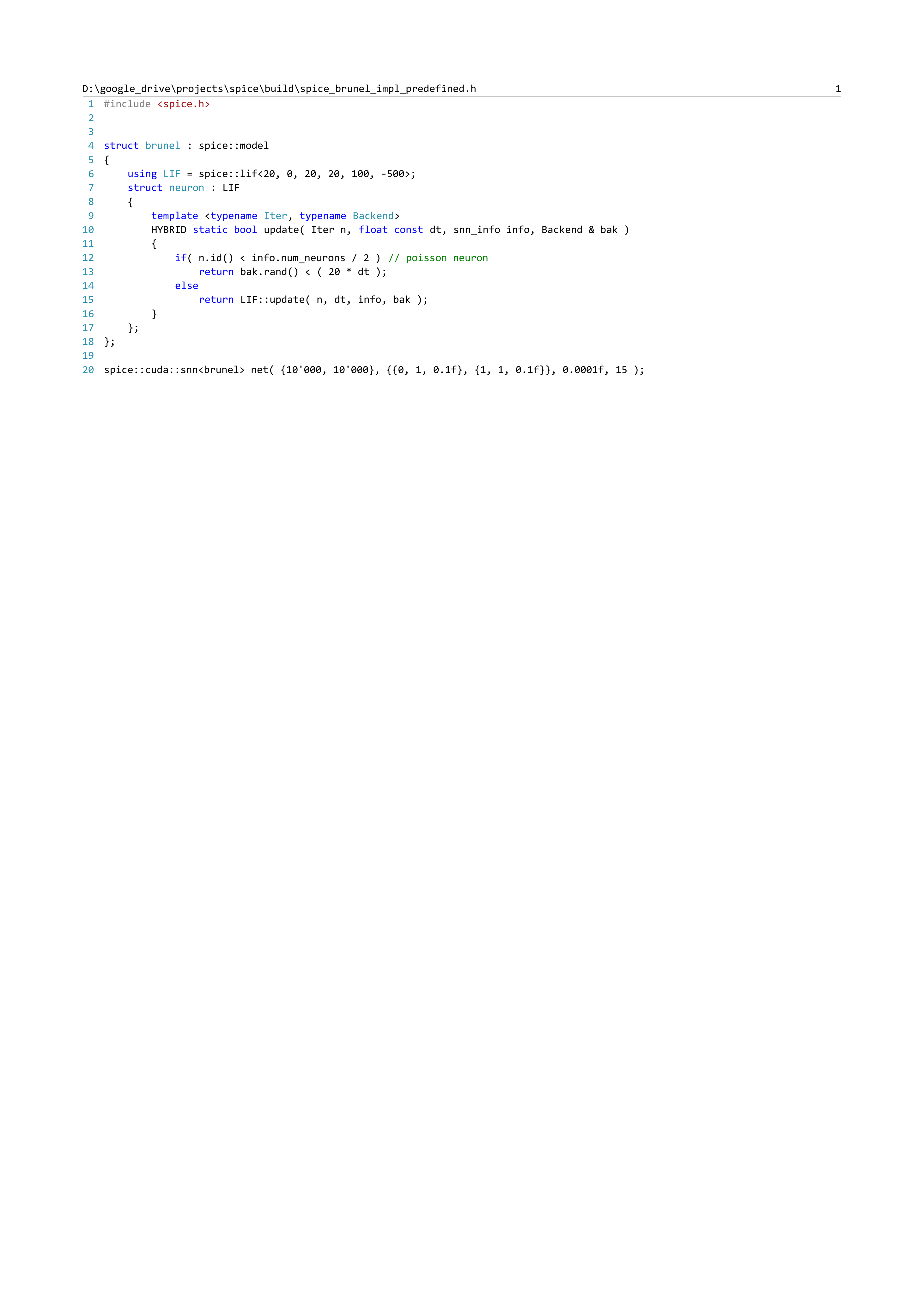}
    \centering
    (a)\hspace{0.2\linewidth}(b)\hspace{0.2\linewidth}(c)\hspace{0.2\linewidth}(d)
    \caption{Comparison of programming effort required to implement the Brunel model in various simulators. Code taken from either official or peer-reviewed samples, excessive comments removed, formatted identically with clang-format. (a) SPIKE (b) GeNN (c) Ours, implemented ex nihilo (d) Ours, implemented using the supplied LIF neuron-building block.}
    \label{figure:loc_comparison}
\end{figure}

GeNN is highly modular and ships with a collection of basic models and building blocks. Implementing a simulation involves a few stages and languages. It amounts to providing the synapse, neuron and plasticity models (using GeNN's domain-specific language (DSL)), running a proprietary compilation step, which brings everything together in a new translation unit of CUDA C++, and linking against said unit from user code. Model, network size and connectivity parameters have to be provided at \textit{compile} time, requiring re-compilation
for any change.
During experimentation we had to maintain several GeNN simulators, for various models and various network sizes, with each one requiring distinct compilation. The user can choose to extend existing models or provide them ex nihilo. Implicit initialization is performed efficiently on the GPU. It is straight-forward to download timestamped spikes.

While designing our simulator, as users of SNN simulation software ourselves, we tended to features that we found to be the most positively impactful in our SNN research, and that will most likely accelerate research for the community, too.
It was highly important to us
that the entire workflow can be supported by a widespread, well-maintained single tool, like the C++ compiler. Everything, from building to using, was far simpler for SPIKE because of this. We also appreciate GeNN's design choice to provide scaffolding for custom models, rather than fixing building blocks in place, while, at the same time, making stable SNN parts like graph construction, spike propagation, delays, etc. opaque to the user.

We push the envelope on that front and accommodate this functionality by replacing GeNN's DSL paradigm with actual C++ code, compiled by and adhering to the same rules as the tool-chain used for the rest of the framework. Both SPIKE and GeNN, to different degrees, offer the means to reuse components. We too provide this option, to an elevated degree. On the one hand, the user can write arbitrary C++ code to finely control simulation stages. On the other hand, a library of standardized functionality is provided too and is made available for the user to employ when necessary. Through our design choice it naturally derives that established reuse patterns within C++ can be used without restriction, too.

We also wanted to make sure that
the required API calls were minimal in count and verbosity. By example (see Fig.~ \ref{figure:loc_comparison}): it can be seen that our simulator requires the least amount of code to bootstrap. Table \ref{table:comp_programming_models} summarizes the qualitative comparison made in this section.

\begin{table}
    \begin{center}
    \caption{Comparison of APIs}
    \label{table:comp_programming_models}
    \begin{tabular}{|l|c|c|c|}
        \hline
        & {\bf SPIKE} & {\bf GeNN} & {\bf Ours} \\
        \hline\hline
        Generality & $\bigstar$ & $\bigstar\bigstar$ & $\bigstar\bigstar\bigstar$ \\
        \hline
        Code Safety & $\bigstar\bigstar\bigstar$ & $\bigstar\bigstar$ & $\bigstar$ \\
        \hline
        Composability & $\bigstar\bigstar$ & $\bigstar\bigstar$ & $\bigstar\bigstar\bigstar$ \\
        \hline
        Reusability & $\bigstar$ & $\bigstar\bigstar$ & $\bigstar\bigstar\bigstar$ \\
        \hline
        Transparency & $\bigstar$ & $\bigstar\bigstar$ & $\bigstar\bigstar\bigstar$ \\
        \hline
        Native Tooling Support & \checkmark & $\times$ & \checkmark \\
        \hline
        Dynamic Parameters & $\bigstar\bigstar\bigstar$ & $\bigstar$ & $\bigstar\bigstar\bigstar$ \\
        \hline
        Ease of Use & $\bigstar\bigstar\bigstar$ & $\bigstar$ & $\bigstar\bigstar\bigstar$ \\
        \hline
    \end{tabular}
    \end{center}
\end{table}
\section{Results}
\label{chapter:results}
We compare our performance with SPIKE and GeNN in a series of benchmarks using adaptions by~\cite{Ahmad2018} of three well-established models: Vogels-Abbott (henceforth referred to as ``Vogels'')\cite{vogels2005}, Brunel, and Brunel with plasticity (referred to as ``Brunel+'')\cite{brunel2000}. All models
\begin{itemize}
    \item use leaky integrate and fire (LIF) neurons.
    \item subdivide neurons into two groups (aka populations): inhibitory and excitatory. Inhibitory neurons have a high potential leak rate, inhibiting overall network activity. Excitatory neurons have a low to zero leak rate, exciting overall network activity.
\end{itemize}
They differ in their stimulation, dynamics, and parameterization. In Vogels, a constant background voltage excites all neurons. In Brunel, a population of Poisson firing-neurons excites the remainder of the network. Finally, Brunel can be run with and without STDP. A detailed overview over both models can be found in~\cite{Ahmad2018}, appendices A \& B.

Vogels and Brunel have been conceived nearly 20 years ago and the network sizes they were originally tuned for (\num{4000} neurons and \num{20000} neurons respectively) are not remotely large enough to stress modern GPUs and accurately compare
SNN simulator performance.
Simply increasing the network size alters both models' firing patterns beyond recognition.
Therefore
we propose a minor change to both models, which
\begin{itemize}
    \item does not alter the models in any way for the original network sizes.
    \item retains the models' characteristic firing patterns for all other network sizes (up to billions of synapses).
\end{itemize}
We do so by scaling the synaptic weights
before they are added to the neuron potentials. Normally, when a neuron receives a spike, the weight $W$ of the synaptic connection over which the spike was received is added to the neuron potential $V$: $V = V + W$, which we change into $V = V + c * W$. $c$ depends on the network size and differs for both models:
\begin{equation}
    c =
    \begin{cases}
        \dfrac{\num{16000000}} {|\Neur|^2}, & \text{Vogels} \\[14pt]
        \dfrac{\num{20000}} {|\Neur|}, & \text{Brunel(+)}
    \end{cases}
\end{equation}
As can be seen, when substituting the original network sizes $c$ becomes 1 and thus has no effect. For larger network sizes, both models retain their characteristic firing patterns. For Vogels, the network's average firing rate (ratio of neurons spiking) remains between half and twice the original rate. For Brunel, the average firing rate remains virtually constant.

The benchmarks were performed on a PC with an Intel Core i7-8700 CPU, 32GB of DDR4 2400 RAM, and an NVIDIA GeForce RTX 2080 Ti GPU.

\begin{figure*}
\begin{tabular}{ccc}
    \includegraphics[width = 0.32\textwidth]{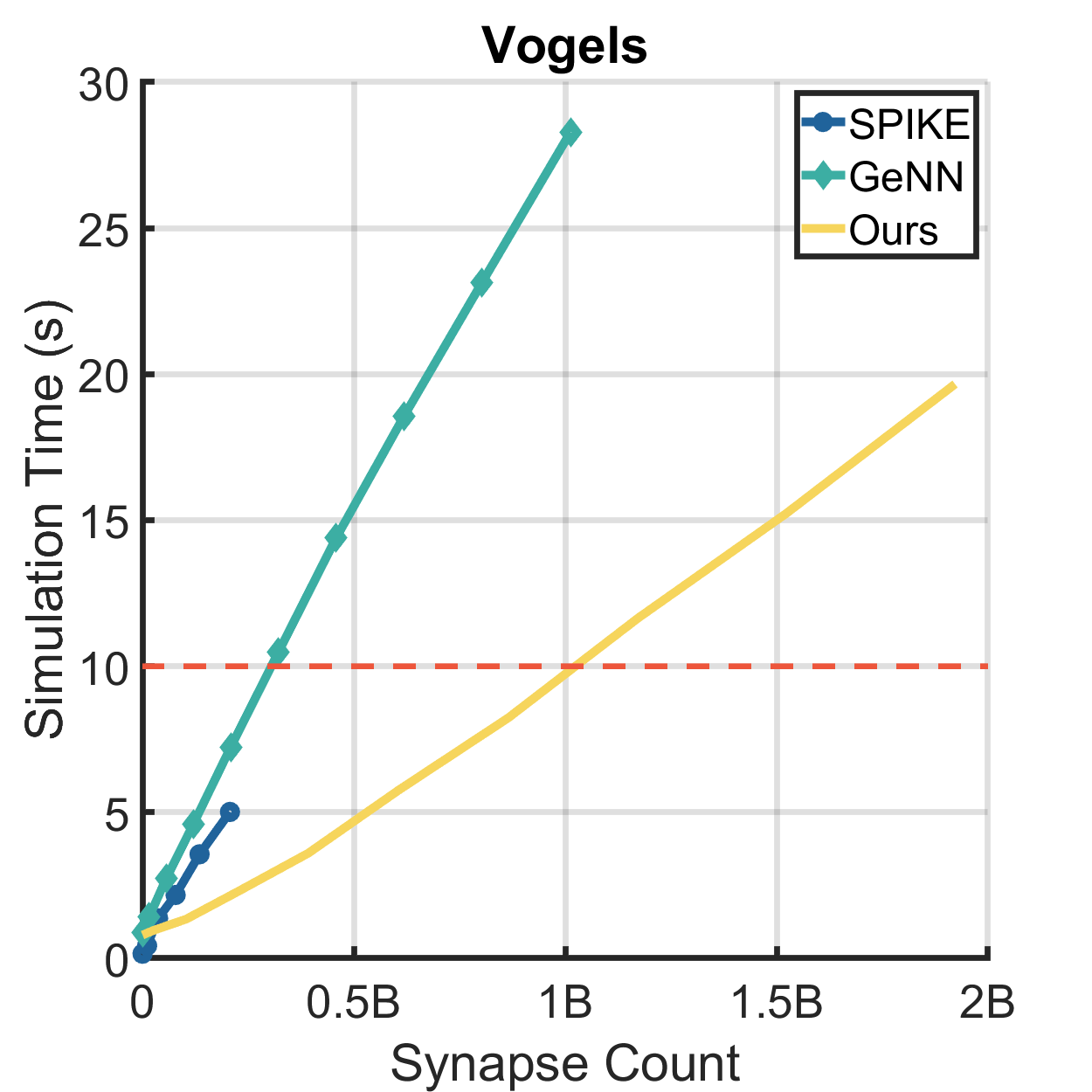} &
    \includegraphics[width = 0.32\textwidth]{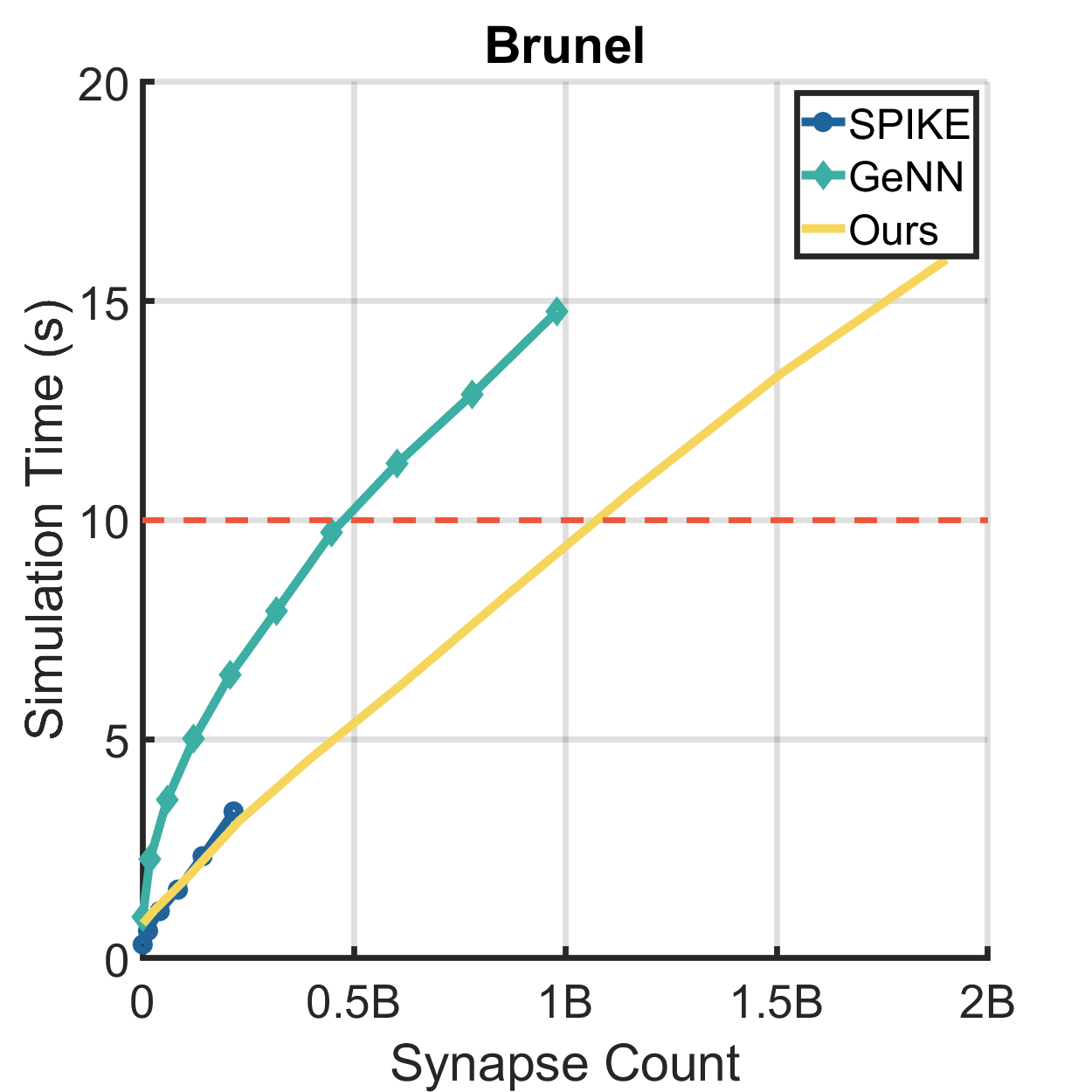} &
    \includegraphics[width = 0.32\textwidth]{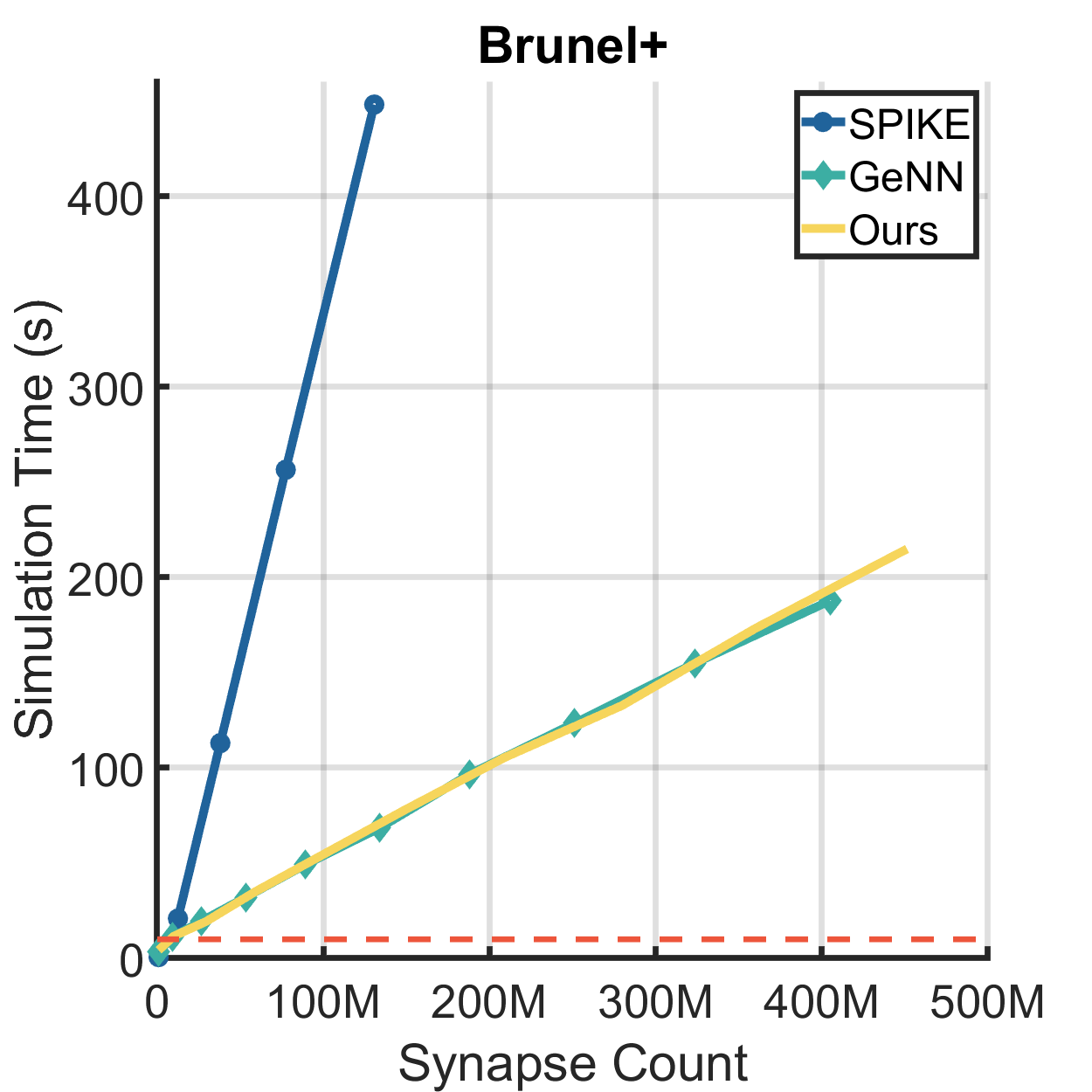}
\end{tabular}
\caption{Simulation time as a function of network size. We vary the synapse count and measure the time in seconds it takes to simulate 10s worth of activity (red dashed line). Left to right: Vogels, Brunel, Brunel+. Graphs are aborted once simulators run out of memory.}
\label{figure:runtime}
\end{figure*}

\begin{figure}
    \centering
    \includegraphics[width=0.36\textwidth]{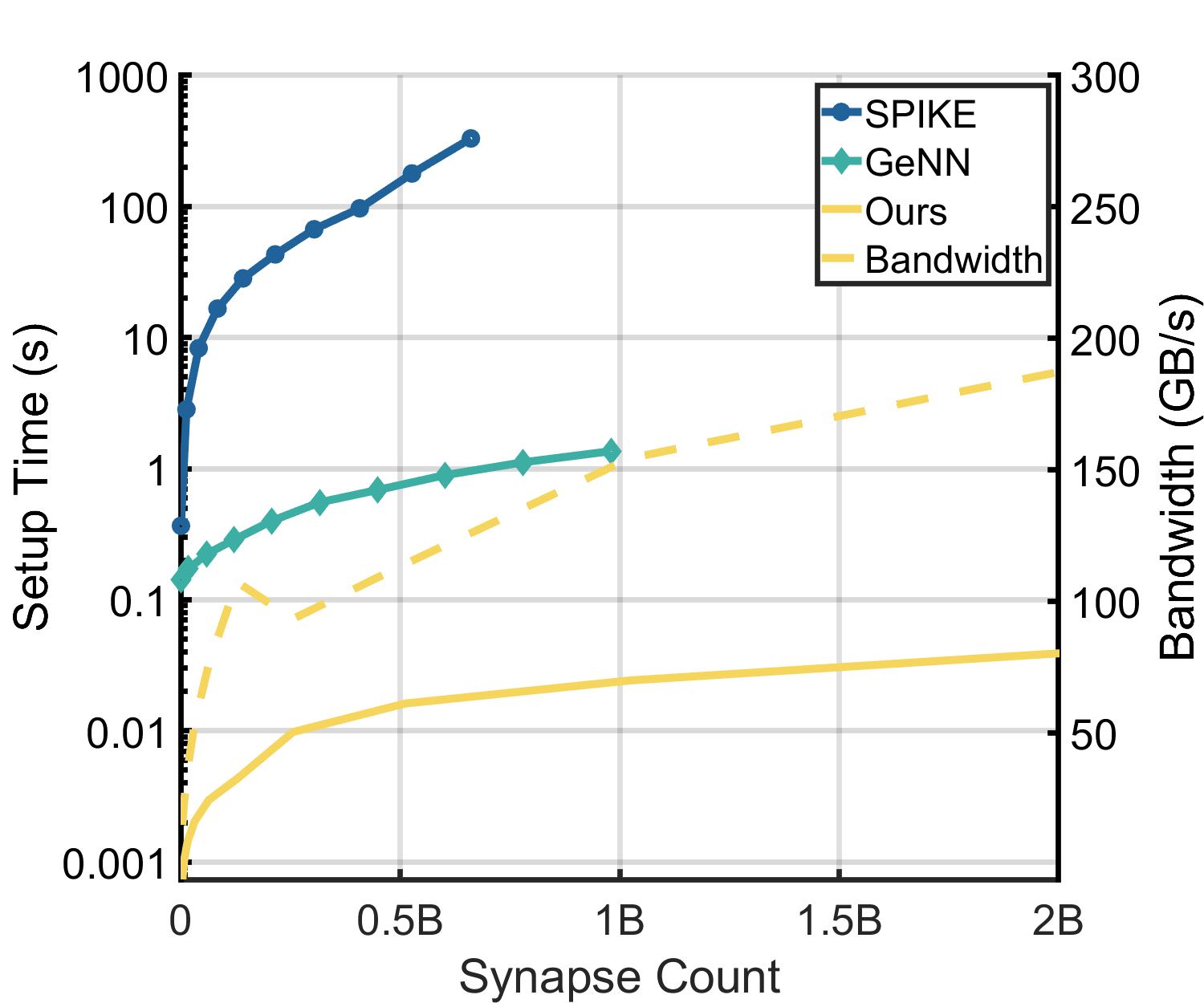}
    \caption{Setup time as a function of network size for \textit{Brunel} model. Y-axis is logarithmic.}
    \label{figure:setuptime}
\end{figure}

\subsection{Benchmarks}
\subsubsection{Simulation time as a function of network size}
We vary the network size (synapse count) and report the average absolute time in seconds it takes to simulate 10 seconds worth of activity for various models (Fig.~\ref{figure:runtime}).

For the tiniest of network sizes, SPIKE is the fastest simulator. However, it is quickly overtaken by us due to its poor scaling. It is also the first simulator to run out of memory, limiting the problem sizes it can be applied to. Compared to GeNN, we are \textasciitilde3x faster for Vogels, \textasciitilde1.5x -- \textasciitilde2x faster for Brunel, and just as fast for Brunel+. We also show near-perfect linear scaling for all models and are the last to run out of memory, allowing us to simulate models with up to double the synapse count compared to our closest competitor.

\subsubsection{Setup time as a function of network size}
One important scenario in SNN simulations is running many experiments back-to-back with different parameters/network sizes. Thus, a fast setup time is desirable to maximize time spent inside simulation. Setup consists of three main steps:
(1)~Constructing the network, (2)~initializing neurons and (3)~initializing synapses.
Steps 2 and 3 are trivial and are really a measure of memory bandwidth rather than algorithmic efficiency.
Synapse state dominates setup time for models where it is present.
The interesting part is step 1 because it is a costly and complex operation. Therefore, we base this benchmark on Brunel which has no synapse state and simple neuron initialization, making its setup time mostly dependent on graph construction, while still being a real-world model. We vary the network size and report the absolute setup time in seconds. We also benchmark and report our simulator's memory bandwidth (Fig.~\ref{figure:setuptime}).

SPIKE initializes on the CPU which is why it is 2 orders of magnitude slower than GeNN which initializes on the GPU.
Compared to GeNN, which allocates memory for every initialization, we only allocate memory when necessary (using hysteresis). When running many experiments back to back we thus often pay the price of allocation only once at the beginning. This, bundled with our highly efficient graph construction algorithm, gives us another 2 orders of magnitude improvement over GeNN. Fig.~\ref{figure:setuptime} clearly illustrates that the notion of ``setup time not mattering because it only happens once'' is a fallacy. In the time it takes SPIKE to initialize a network with 500 million synapses, we can already simulate 6 minutes worth of activity.

\begin{table}[b]
    \begin{center}
    \caption{Detailed breakdown of our simulator's memory consumption}
    \label{table:memory}
    \begin{tabular}{|l|l|r|r|r|}
        \hline
        \multicolumn{2}{|c|}{{\bf Model}} & {\bf Vogels} & {\bf Brunel} & {\bf Brunel+} \\
        \hline\hline
        \multirow{6}{*}{per neuron} & fields & 16B & 8B & 8B \\
        & spikes & 32B & 60B & 60B \\
        & bitmasks & - & - & 6.25B \\
        & ages & - & - & 4B \\
        & expirations & - & - & 4B \\
        \cline{2-5}
        & total & 48B & 68B & 82.25B \\
        \hline
        \multirow{3}{*}{per synapse} & adjacency list & 4B & 4B & 4B \\
        & fields & - & - & 12B \\
        \cline{2-5}
        & total & 4B & 4B & 16B \\
        \hline
    \end{tabular}
    \end{center}
\end{table}

\subsubsection{Memory consumption as a function of network size}
We report \textit{our} simulator's memory consumption in gigabytes. It becomes apparent that it is entirely dominated by synapse count. For a detailed breakdown of the total memory consumption see Table~\ref{table:memory}.

\begin{figure}
    \centering
    \includegraphics[width=0.36\textwidth]{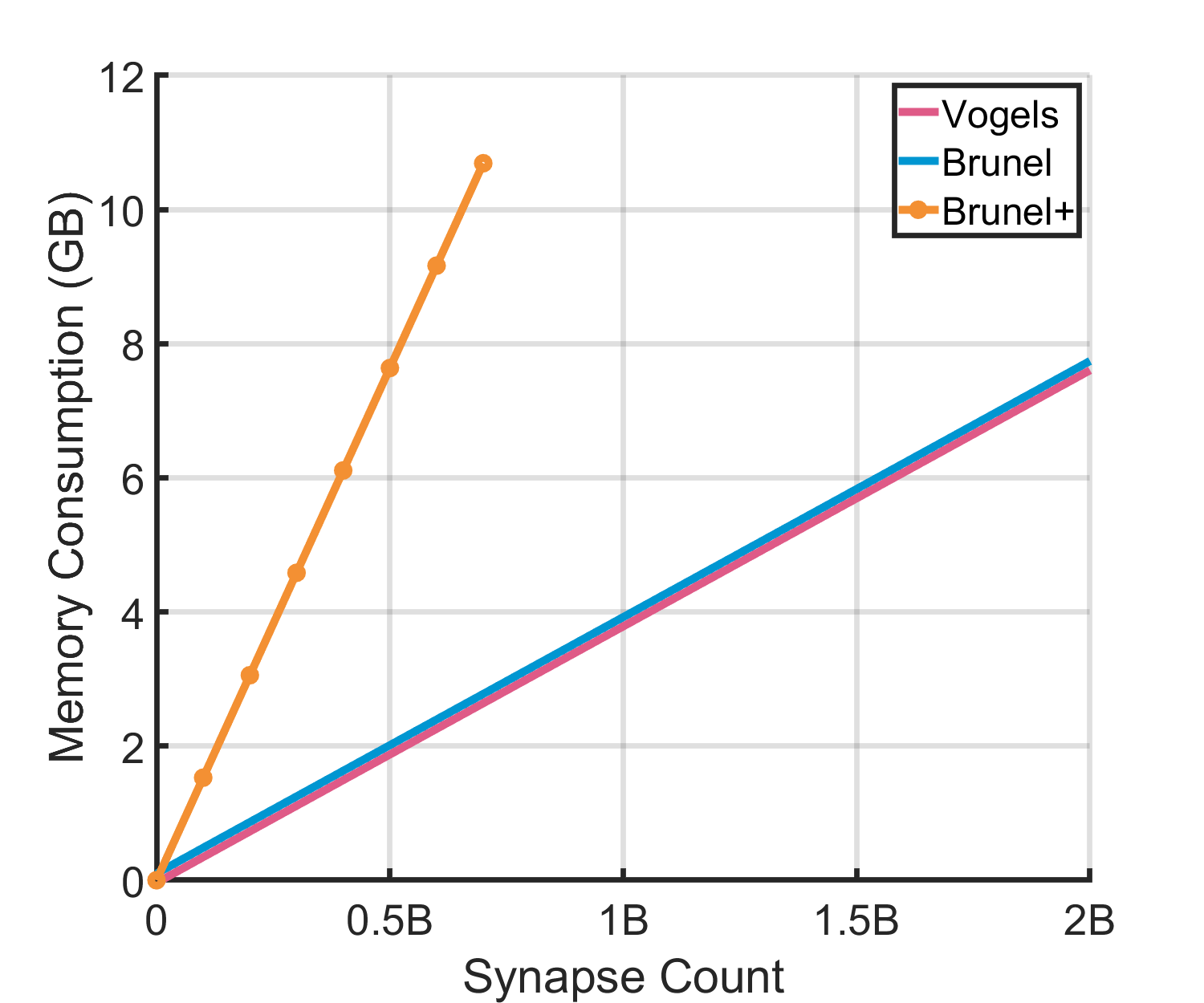}
    \caption{\textit{Our} simulator's memory consumption for various models as a function of network size.}
    \label{figure:memory}
\end{figure}


\section{Conclusions \& Future Work}
We presented a SNN simulator which is faster and consumes less memory than the state of the art, allows the specification of more general models, offers a simpler and less verbose API and build process.
Spiking Neural Networks regained a lot of traction and popularity recently. In spite of simulation improving by two orders of magnitude in the last year alone, SNNs still have a long way to go in order to compete with conventional ANNs. There is one virtually unexplored optimization in SNN simulation: Mutli-GPU parallelization. A promising approach might be to parallelize simulation across neuron populations, similarly to how PipeDream \cite{pipedream2018} parallelizes backpropagation across layers.

\section*{Acknowledgements}
This work was partially supported by the European Community through the project Co4Robots (H2020-731869).

\bibliographystyle{IEEEtran.bst}
\bibliography{spiking}

\end{document}